\documentclass[letterpaper]{article} 
\usepackage{aaai2026}  
\usepackage{times}  
\usepackage{helvet}  
\usepackage{courier}  
\usepackage[hyphens]{url}  
\usepackage{graphicx} 
\usepackage{natbib}  
\usepackage{caption} 
\usepackage{algorithm}
\usepackage{algorithmic}

\usepackage[utf8]{inputenc}
\usepackage[T1]{fontenc}    
\usepackage{url}            
\usepackage{booktabs}       
\usepackage{amsfonts}       
\usepackage{nicefrac}       
\usepackage{microtype}      
\usepackage{xcolor}         
\usepackage{multirow}
\usepackage{graphicx}       

\usepackage{lipsum} 
\usepackage{enumitem}
\usepackage{amsmath}

\usepackage{amsthm}

\usepackage{amsmath}
\usepackage{todonotes}
\usepackage{threeparttable}
\usepackage{newfloat}
\usepackage{listings}
\DeclareCaptionStyle{ruled}{labelfont=normalfont,labelsep=colon,strut=off} 
\floatstyle{ruled}
\newfloat{listing}{tb}{lst}{}
\floatname{listing}{Listing}
\title{TermGPT: Multi-Level Contrastive Fine-Tuning for Terminology Adaptation in Legal and Financial Domains}
\author{
    Yidan Sun\textsuperscript{\rm 1}, Mengying Zhu\textsuperscript{\rm 1}\footnote{Corresponding author}, Feiyue Chen\textsuperscript{\rm 1}, Yangyang Wu\textsuperscript{\rm 1}, Xiaolei Dan\textsuperscript{\rm 2}, Mengyuan Yang\textsuperscript{\rm 1}, Xiaolin Zheng\textsuperscript{\rm 1}, Shenglin Ben\textsuperscript{\rm 1}\\
}
\affiliations{
    \textsuperscript{\rm 1}Zhejiang University, China\\
    \textsuperscript{\rm 2}	Innovation Department, National FinTech Risk Monitoring Center, China \\

    \{yidansun, mengyingzhu, fychen02, zjuwuyy, yangmy412, xlzheng, benshenglin\}@zju.edu.cn \\
    danxiaolei@nfrm.org.cn \\
}

\usepackage{bibentry}

\begin{document}

\maketitle

\begin{abstract}

Large language models (LLMs) have demonstrated impressive performance in text generation tasks; however, their embedding spaces often suffer from the isotropy problem, resulting in poor discrimination of domain-specific terminology, particularly in legal and financial contexts. This weakness in terminology-level representation can severely hinder downstream tasks such as legal judgment prediction or financial risk analysis, where subtle semantic distinctions are critical. 
To address this problem, we propose \texttt{TermGPT}, a multi-level contrastive fine-tuning framework designed for terminology adaptation. We first construct a sentence graph to capture semantic and structural relations, and generate semantically consistent yet discriminative positive and negative samples based on contextual and topological cues. We then devise a multi-level contrastive learning approach at both the sentence and token levels, enhancing global contextual understanding and fine-grained terminology discrimination. 
To support robust evaluation, we construct the first financial terminology dataset derived from official regulatory documents. Experiments show that \texttt{TermGPT} outperforms existing baselines in term discrimination tasks within the finance and legal domains.
\end{abstract}


\begin{links}
    \link{Code}{https://github.com/Thoams0211/TermGPT}
\end{links}

\section{Introduction}

Large language models (LLMs) have made significant progress in text generative tasks. However, the embedding spaces learned from LLMs often suffer from the isotropy problem, where the token embeddings are distributed too uniformly in high-dimensional space, resulting in insufficient semantic discriminability \cite{mickus2024isotropy,mu2017all,tsukagoshi2025redundancy}. This problem hampers the LLMs’ ability to distinguish subtle yet crucial semantic differences between domain-specific terminology, thereby limiting accurate understanding of terminology.

\begin{figure}[t]
    \centering  \includegraphics[width=0.95\linewidth]{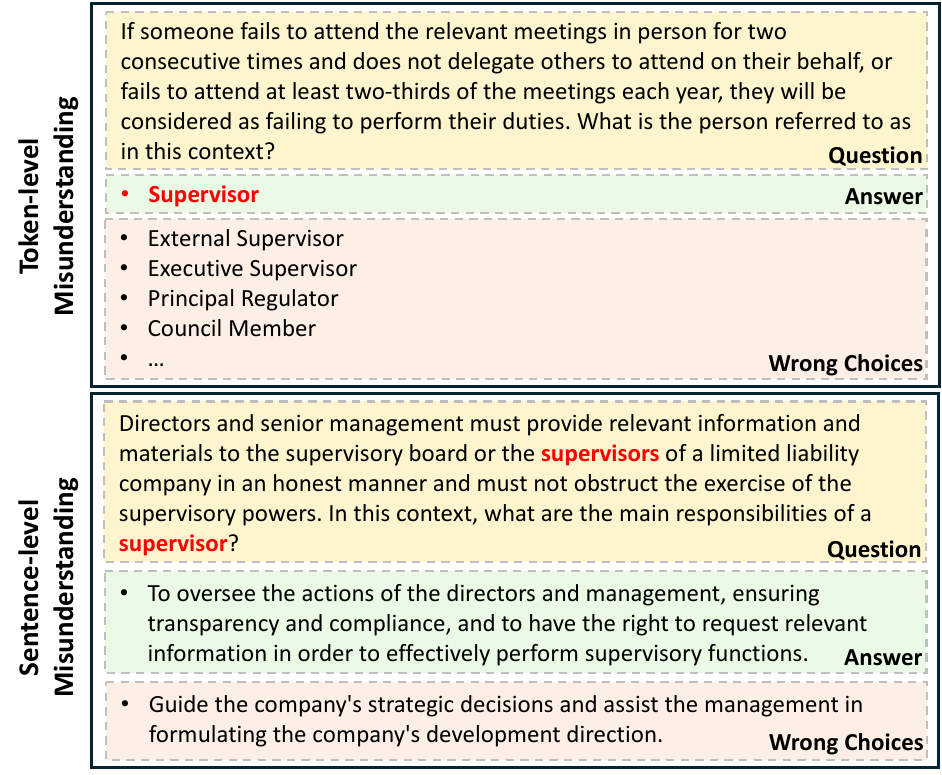}
    \caption{Motivation Example: Terminology Misunderstanding in a Loan Application.}
    \label{fig:case_study}
\end{figure}

This limitation is particularly problematic in high-stakes domains, such as finance and law, where precise interpretation of terminology is critical for downstream tasks, including loan application and compliance advisory services \cite{chen2023disc, liu2023fingpt}. 
In these scenarios, even slight confusion between related terms may lead to incorrect reasoning and serious real-world consequences.
Taking loan application as an example, as shown in Figure \ref{fig:case_study}, an LLM deployed in a banking system misinterprets term "\textit{supervisor}" as "\textit{executive supervisor}" during a loan review.
This misunderstanding causes the LLM to overlook a key omission of disclosure to the supervisory board. 
As a result, loan is mistakenly approved, exposing institution to financial loss, regulatory penalties, and reputational risk.
To mitigate such risks, there is growing need for fine-tuning methods that enhance the ability of LLMs to distinguish domain-specific terminology with high semantic precision.
In response to this need, we formally define a new task, \textbf{terminology-aware fine-tuning task}, which focuses on improving the representation and discrimination of specialized terms in domain-specific contexts.

Recently, several studies \cite{kim2024hil, rudman2023stable} have focused on improving the quality of token embeddings, demonstrating potential for completing terminology-aware fine-tuning task.
First, the supervised fine-tuning (SFT) method \cite{kim2024efficient} aligns LLMs' output with labeled instruction–response pairs, helping the model distinguish between semantically correct and incorrect responses, and improving separability of token embeddings.
Second, sentence-level contrastive learning methods \cite{gunel2020supervised, lee2020contrastive} fine-tune LLMs by constructing positive and negative pairs at the sentence level, enhancing model’s ability to encode global contextual semantics.
Thirdly, token-level contrastive learning methods \cite{zhang2022frequency,jiang2022simple} apply contrastive objectives at token level, often incorporating frequency-based weighting or regularization to enhance term-level discrimination. 
While these methods offer valuable components for terminology-aware fine-tuning, they are not directly applicable in practice.

This is primarily due to three challenges that arise in the context of terminology-aware fine-tuning.
First, \textit{supervision signals for terminology are ambiguous} \textbf{(CH1)}. Terminology is highly context-dependent, with the same terminology potentially carrying different meanings across legal, financial, or regulatory contexts. This ambiguity makes it difficult to provide consistent supervision during fine-tuning.
Secondly, \textit{terminology is extremely sparse in domain-specific corpora} \textbf{(CH2)}. Terms constitute only 0.08\% of the JecQA dataset \cite{zhong2020jec}. This extreme sparsity leads to weak signal strength, causing term representations to be overwhelmed by general contextual patterns.
Thirdly, \textit{terminology-aware fine-tuning involves critical trade-off in leveraging contextual information} \textbf{(CH3)}. 
Over-reliance on context may obscure term-specific semantics, while ignoring context may isolate terms from their functional usage. Without a proper balance between global context and local term meaning, model struggles to learn accurate and discriminative term representations.

To address the above challenges, we propose \texttt{TermGPT}, a multi-level contrastive learning framework for terminology-aware fine-tuning. Specifically, we construct a sentence graph that jointly captures semantic and structural relationships among sentences. Based on this graph, we introduce a graph-driven data augmentation method that generates more accurate and diverse positive and negative pairs (\textit{for addressing} \textbf{CH1}). We then design a multi-level contrastive learning mechanism.  On the one hand, we propose a token-level contrastive learning mechanism that constructs positive and negative samples around key tokens, thereby preventing important terms from being diluted during training (\textit{for addressing} \textbf{CH2}). On the other hand, we propose a sentence-level contrastive learning mechanism, where context-enriched sentence embeddings are optimized to capture the global semantics of terms, thereby alleviating the limitations of isolated token-level modeling (\textit{for addressing} \textbf{CH3}).

In summary, our main contributions are as follows:
\begin{itemize}
    \item \textit{New task:} 
    We define terminology-aware fine-tuning as a novel task that aims to enhance LLMs’ fine-grained understanding of domain-specific terminology. To the best of our knowledge, this is the first work to formalize terminology-aware fine-tuning as a standalone objective, providing clear benefits for terminology-sensitive downstream tasks.
    \item \textit{Novel framework:} We propose \texttt{TermGPT}, a multi-level contrastive fine-tuning framework that jointly models global context and local term semantics. It enhances representation of sparse and imbalanced terminology while effectively integrating contextual information to improve semantic discrimination.
    \item \textit{Specialized dataset:} We construct a new dataset for financial terminology-aware fine-tuning, consisting of 3,647 domain-specific terms extracted from 425 authoritative financial regulatory documents.
    \item \textit{Extensive experiments:} Experiments on both the newly collected financial dataset and the legal-domain JecQA benchmark show that \texttt{TermGPT} outperforms existing baselines, achieving an average improvement of 6.14\% in terminology Question-Answering (QA) tasks and 2.60\% in terminology Question-Choice Answering (QCA) tasks.
\end{itemize}

\section{Related Work}
Improving the quality of token embeddings has great potential for terminology-aware fine-tuning, as it significantly enhances LLM’s ability to understand and distinguish terms. Therefore, we introduce methods aimed at enhancing the effectiveness of token embeddings.

\textbf{Traditional Embedding Methods.}
Early embedding methods, such as Sentence-Bert \cite{reimers2019sentence}, GTR \cite{ni2021large}, and bge-M3 \cite{chen2024bge}, are based on BERT \cite{devlin2019bert} and T5 \cite{raffel2020exploring} and fine-tuned on large-scale supervised datasets for tasks like retrieval. As LLMs evolved, generative embedding methods like GenEOL \cite{thirukovalluru2024geneol}, MetaEOL \cite{lei2024meta}, and Echo embedding \cite{springer2024repetition} emerged, improving sentence embeddings through diverse transformations and contextualized generation. However, they still struggle with subtle semantic variations, limiting embedding precision and discriminability.

\textbf{Contrastive Learning-based Embedding Methods.} Contrastive learning is widely used in text embedding to align similar texts and separate dissimilar ones. Existing studies can be divided into sentence-level and token-level mechanisms.
On the one hand, \textbf{sentence-level contrastive learning} builds positive/negative pairs and optimizes embeddings with methods like InfoNCE \cite{oord2018representation} and SimCSE \cite{gao2021simcse}. Approaches such as GritLM \cite{muennighoff2024generative}, LLM2Vec \cite{behnamghader2024llm2vec}, AutoRegEMbed \cite{deng2025following} and NV-embed \cite{lee2024nv} enhance sentence embedding by introducing bidirectional or latent attention to capture global semantics. Nevertheless, sentence-level contrastive learning often dilutes token-level semantics, reducing effective for terminology-sensitive tasks.
On the other hand, \textbf{token-level contrastive learning} enhances terminology distinction by optimizing individual tokens. CT \cite{jiang2022simple} introduced contrastive token learning to improve terminology discrimination. TCL and FCL \cite{zhang2022frequency} increased the weight of low-frequency tokens to improve representation learning. SimCTG \cite{su2022contrastive} penalized token similarity to learn more distinguishable embeddings, boosting the recognition and generation of fine-grained semantic differences. However, token-level contrastive learning focuses on isolated token optimization, limiting the capture of global semantics and interdependencies, and hindering sentence embedding.
\begin{figure*}[t]
    \centering
    \includegraphics[width=\linewidth]{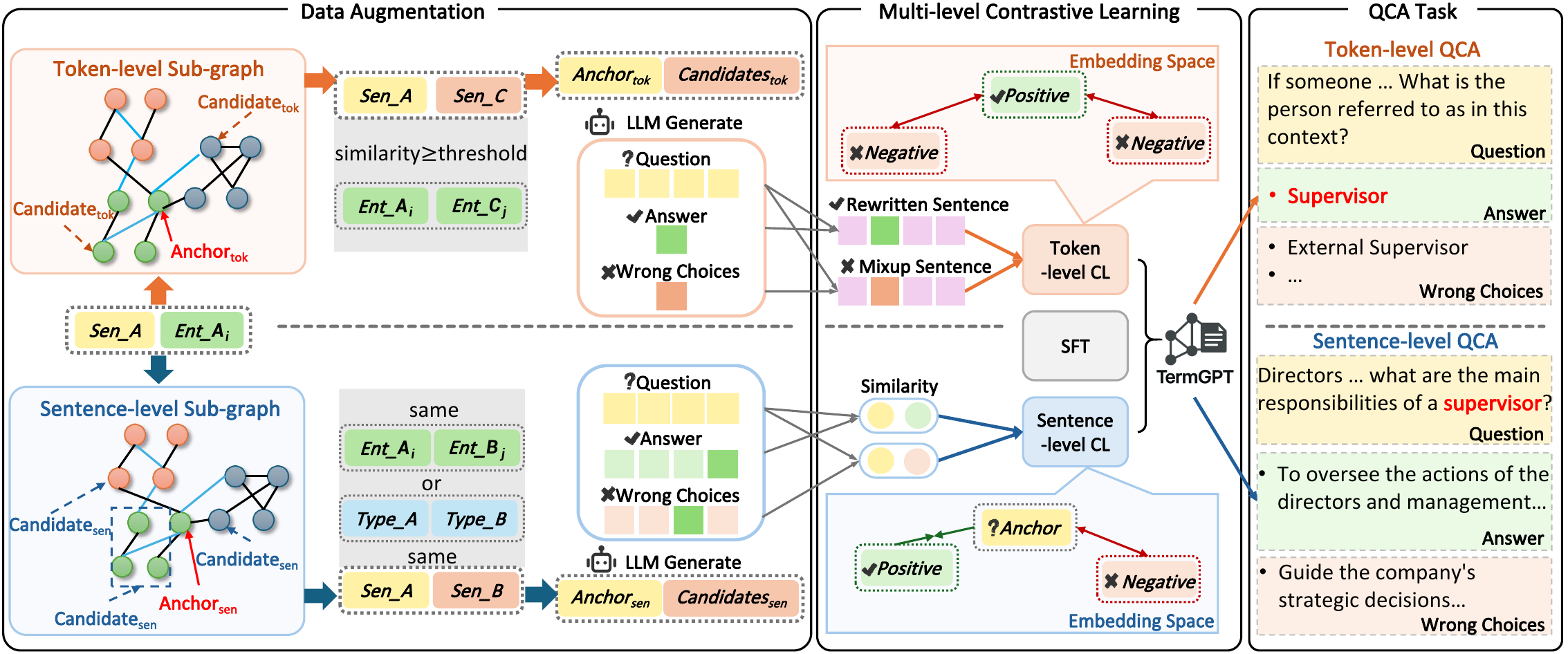}
    \caption{Overall framework of \texttt{TermGPT}. We first construct a sentence graph with sentences as nodes and different semantic and structural relationships as edges, where edges representing semantic ambiguity are black and lexical ambiguity edges are blue. Each node is used as an anchor sample, and its candidate samples are used for data augmentation to generate QCA pairs. Finally, contrastive learning is applied at different levels to distinguish the differences in terminology embeddings based on the QCA categories.}
    \label{fig:model}
\end{figure*}
\section{Preliminary}
In this section, we will introduce basic concepts of terminology-aware fine-tuning, the QCA task, and sentence graph construction.

\textbf{Terminology-aware Fine-tuning.} Terminology-aware fine-tuning reduces the probability of terminology misuse by minimizing the cross-entropy loss. Given a question \( q \), the correct answer \( a \), and candidate incorrect answers \( c_1, c_2, \dots, c_k \), the model computes the probability \( P_{\theta}(a|q) \) and is trained to maximize it by minimizing the loss \( -\sum_{i=1}^{n} \log P_{\theta}(a_i | q_i) \). This enables LLMs to more accurately distinguish between semantically similar but distinct terms, significantly reducing the risk of misuse in domain-specific applications.

\textbf{The QCA Task.} In the QCA task, the objective is to select the correct answer from multiple candidate choices \( a_i, c_{i_1}, c_{i_2}, c_{i_3} \), given the question \( q_i \). Specifically, the input sample \( Q_{{sen}} = (q_i, a_i, c_{i1}, c_{i2}, c_{i3}) \) contains the question and its candidate answers, and the LLM is required to understand the context of the question \( q_i \) and correctly select \( a_i \) from the candidates.

\textbf{Sentence Graph.} In the sentence graph \( G = (V, E) \), nodes represent sentences and edges represent semantic and structural relationships between sentences. There are various semantic relationships between sentences, including sentences that contain the same entities, which are often prone to confusion at the sentence-level semantics. Sentences containing similar entities are more likely to cause confusion at lexical meaning level. Additionally, some sentences may not contain exactly the same entities but may belong to the same semantic category, such as legal clauses in the same field. These sentences of the same type often exhibit high similarity in terminology of theme or structure, which can still lead to misjudgment at sentence meaning level. Therefore, in the sentence graph \( G \), different types of confusion are represented by different edges. Specifically, for a sentence \( s \), we perform entity extraction by combining LLM with a schema, obtaining an entity set \( C = (c_1, c_2, \dots, c_k) \). If two sentences \( s_i, s_j \) contain the same entities or have the same type, they are connected with an edge denoted as \( \text{edge}_{\text{sen}}(s_i, s_j) \). While, if two sentences \( (s_i, s_j) \) contain similar entities \( (c_i, c_j) \), and the embeddings of \( c_i \) and \( c_j \) exceed a certain threshold \( \theta \), an edge \( \text{edge}_{\text{token}}(s_i, s_j) \) is connected between the two sentences.

\section{Methodology}
In this section, we introduce the \texttt{TermGPT} framework, which consists of a data augmentation strategy and a multi-level contrastive learning mechanism. We begin by describing the data augmentation method based on a sentence graph. Then, we present the multi-level contrastive learning mechanism from both sentence-level and token-level perspectives. Finally, we summarize the overall optimization pipeline.

\subsection{Data Augmentation}

In the terminology-aware fine-tuning dataset, terms typically occupy only a small portion of the dataset, which makes the embedding of terminology susceptible to dilution in sentence-level modeling, thereby affecting the LLM's ability to perceive fine-grained differences between terms. To effectively enhance the representational ability of terminology, we propose a sentence graph-based data augmentation method. By constructing a graph that captures semantic relationships between sentences, we generate more diverse and informative positive and negative sample pairs, thus improving the model's discriminative ability. The edges in the sentence graph can reflect various semantic relations, helping the model capture the diversity and subtle differences between terms. This method can effectively enrich the dataset, thereby enhancing the semantic differentiation ability of terminology and improving the model's reasoning performance.

Specifically, in the sentence graph, we first traverse each node, with the current selected node being the anchor node \( s_{\text{anchor}} \). For each node \( s_j \) connected to it by an \( edge_{\text{sen}} \), if the cosine similarity between its sentence-level embedding vector \( \mathbf{e}_{\text{sen}}(s_j) \) and the anchor node's sentence-level embedding vector \( \mathbf{e}_{\text{sen}}(s_{\text{anchor}}) \) is greater than a threshold \( \theta_{\text{sen}} \), then \( s_j \) is added to the candidate set \( S_{\text{sen}}(s_{\text{anchor}}) \) of \( s_{\text{anchor}} \). For each node \( s_i \) connected to \(s_{anchor}\) by an \( edge_{\text{tok}} \), let the similar entities be \( c_{\text{anchor}} \) and \( c_i \). Then \( c_i \) is directly added to the candidate set \( C_{\text{tok}}(s_{\text{anchor}}, c_{\text{anchor}}) \) of \( s_{\text{anchor}} \).

For any candidate set \( S_{\text{sen}}(s_{\text{anchor}}) \), we have the LLM generate a question \( q \) with \( s_{\text{anchor}} \) as anchor, and \( s_{\text{anchor}} \) as the correct option \( a \), \( s_j \in S_{\text{sen}}(s_{\text{anchor}}) \) as the incorrect option. Additionally, two hard negative samples \( s_{j_1} \) and \( s_{j_2} \) are generated. This set of QCA samples \( (q, a, s_j, s_{j_1}, s_{j_2}) \) is added to the dataset \( Q_{\text{sen}} \). Similarly, for any candidate set \( C_{\text{tok}}(s_{\text{anchor}}, c_{\text{anchor}}) = \{ c_1, c_2, \dots, c_k \} \), we have the LLM generate a question \( q \) based on \( s_{\text{anchor}} \) for \( c_{\text{anchor}} \), with \( c_{\text{anchor}} \) as the correct option \( a \) and \( \{ c_1, c_2, \dots, c_k \} \) as incorrect options. This set of QCA samples \( (q, a, c_1, \dots, c_k) \) is then added to the dataset \( Q_{\text{tok}} \).

\subsection{Multi-level Contrastive Learning}
Terminology-aware fine-tuning involves a trade-off between contextual understanding and terminology-specific semantics. 
To balance this trade-off, we propose a multi-level contrastive learning mechanism that integrates both global context and local semantic features for learning accurate terminology embeddings.

\noindent \textbf{Sentence-level contrastive learning.} We first utilize a LLM to generate embeddings for the input sample $x_i \in Q_{\text{sen}} = (q_i, a_i, c_{i1}, c_{i2}, c_{i3})$, where $q_i$ is the question, $a_i$ is the correct answer, and $c_{i1}, c_{i2}, c_{i3}$ are the three candidate incorrect answers. We then activate a bidirectional attention mechanism to ensure the full integration of context information. Next, we optimize the contrastive learning objective using the InfoNCE loss function by computing the similarity between the embeddings of the question and the correct answer, as well as the similarity between the question and the three candidate incorrect answers. The loss function $L_{\text{sen}}$ is defined as:

\begin{equation}
    L_{\text{sen}} = \sum_{x_i \sim Q_{\text{sen}}} -\log \frac{\exp(\mathbf{e}_{q_i} \cdot \mathbf{e}_{a_i} / \tau)}{\sum_{j=1}^{4} \exp(\mathbf{e}_{q_i} \cdot \mathbf{e}_{c_{ij}} / \tau)},
\end{equation}
where $\tau$ is the temperature parameter. The goal is to maximize the similarity between the question and the correct answer, while minimizing the similarity between the question and the incorrect options, thus enhancing the LLM’s ability to discriminate the semantics of the terminology.

\noindent \textbf{Token-level contrastive learning.} Given the input sample $x_i \in Q_{\text{tok}} = (q_i, a_i, c_{i1}, \dots, c_{ik})$, we use the LLM to convert $(q_i, a_i)$ into a declarative sample $t_i$, where $a_i$ appears as a subsequence $\{ a_i \} \subset t_i$ within the sentence $t_i$, and the set of negative samples is denoted as $C_{\text{neg}} = (c_{i1}, \dots, c_{ik})$. We treat the correct answer $a_i$ as a positive sample $z^+$, and each incorrect answer $c_{ij}$ as a negative sample $z^-$, and mix the positive and negative sample pair $(z^+, z^-)$ with the sentence $t$ to form a mixed sequence $\tilde{t} = \text{Mix}(t, z^+, z^-)$. During the mixing process, the function replaces $z^+$ in the sentence $t$ with $z^-$. We define a sequence $\varphi$ with the same length as $\tilde{t}$, where 1 indicates tokens from $z^+$ or $t$, and 0 indicates tokens from $z^-$. Based on the mixed sequence $\tilde{t}$ and $\varphi$, we design the following loss function $l_{\text{mix}}$:

\begin{align}
&l_{\text{mix}}(t, z^+, z^-, q) = - \sum_{j=1}^{|\tilde{t}|} \Big[ \varphi_{i,j} \log p_\theta(\tilde{t}_{i,j} | \tilde{t}_{i,<j}, q) \nonumber \\
& \quad + (1 - \varphi_{i,j}) \log \left(1 - p_\theta(\tilde{t}_{i,j}|\tilde{t}_{i,<j}, q)\right) \Big].
\end{align}
For the entire set $Q_{\text{tok}}$, the token-level contrastive learning loss function $L_{\text{tok}}$ is expressed as:

\begin{equation}
    L_{\text{tok}} = \sum_{x \sim Q_{\text{tok}}} \sum_{i=1,\dots,k} l_{\text{mix}}(t, z^+, z^-_i, q).
\end{equation}
By integrating sentence-level and token-level contrastive learning, we construct a robust framework that enhances both fine-grained terminology differentiation and contextual coherence, thereby improving overall model performance.

\begin{algorithm}[ht]
\caption{Optimization Pipeline of \texttt{TermGPT}}
\label{alg:opt}
\begin{algorithmic}
\STATE \textbf{Input:} $Q_{\text{sen}}, Q_{\text{tok}}$
\STATE \textbf{Output:} $\theta$
\STATE \textit{// Step 1: Supervised Fine-Tuning (SFT)}
\FOR{$x \in Q_{\text{sen}} \cup Q_{\text{tok}}$}
    \STATE Compute $L_{\text{SFT}}$.
    \STATE Update $\theta$ to minimize $L_{\text{SFT}}$.
\ENDFOR
\STATE \textit{// Step 2: Sentence-level Contrastive Learning}
\FOR{$(q_i, a_i, c_{i_1}, c_{i_2}, c_{i_3}) \in Q_{\text{sen}}$}
    \STATE Let $\mathbf{e}_{q_i}, \mathbf{e}_{a_i}, \mathbf{e}_{c_{i1}}, \mathbf{e}_{c_{i2}}, \mathbf{e}_{c_{i3}}$ be the embeddings.
    \STATE Compute $L_{sen}$.
    \STATE Update $\theta$ to minimize $L_{sen}$.
\ENDFOR
\STATE \textit{// Step 3: Token-level Contrastive Learning}
\FOR{$(q_i, a_i, c_{i_1}, \dots, c_{i_k}) \in Q_{\text{tok}}$}
    \STATE Convert $(q_i, a_i)$ to a declarative sentence $t_i$ using the LLM.
    \STATE Let $\tilde{t} = \text{Mix}(t, z^+, z^-)$ be the mixed sequence.
    \STATE Let $\mathbf{e}_t$ be the token embeddings of $\tilde{t}$ and $\varphi$ be the binary sequence indicating whether a token is from $z^+$ or $z^-$.
    \STATE Compute $l_{\text{mix}}(t, z^+, z^-, q)$.
    \STATE Compute $L_{\text{tok}} = \sum_{x \sim Q_{\text{tok}}} \sum^{i=1,\dots,k}_{z^-_i \sim C_{\text{Neg}}} l_{\text{mix}}(t, z^+, z^-_i, q)$.
    \STATE Update $\theta$ to minimize $L_{tok}$.
\ENDFOR
\end{algorithmic}
\end{algorithm}
\subsection{Optimization Pipeline}

In the overall optimization pipeline, we first perform a unified SFT on the sentence set \( Q_{\text{sen}} \) and token set \( Q_{\text{tok}} \). During this phase, the LLM learns the mapping from the original input \( q_i \) to the target output \( y_i \) by fine-tuning on downstream tasks. The loss function for SFT, denoted as \( L_{\text{SFT}} \), is defined as:

\begin{equation}
    L_{\text{SFT}} = - \sum_{x \in Q_{\text{sen}} \cup Q_{\text{tok}}} \log p_\theta(y_q | x_q),
\end{equation}
where \( x_q \) represents the input sample, \( y_q \) is the corresponding target output. After completing the SFT phase, sentence-level contrastive learning is first applied to \( Q_{\text{sen}} \), maximizing the similarity between the question and the correct answer while minimizing the similarity between the question and the incorrect options. Subsequently, \texttt{TermGPT} performs token-level contrastive learning on \( Q_{\text{tok}} \), further refining the fine-grained distinction of terminology through mixed sequences \( \tilde{t} \) and \( \varphi \). The pseudocode is as Algorithm \ref{alg:opt}.

\section{Experiment}
\label{sec:experiment}

In this section, we present extensive experiments to answer the following questions: 

\textbf{RQ1:} How does \texttt{TermGPT} perform compared to the baselines on the terminology QCA tasks?

\textbf{RQ2:} How does \texttt{TermGPT} perform compared to the baselines on the terminology QA tasks?

\textbf{RQ3:} How does the multi-level contrastive learning mechanism affect \texttt{TermGPT}’s ability to understand domain-specific terminology?

\textbf{RQ4:} How well does \texttt{TermGPT} generalize to terminology understanding tasks across different domains?

\subsection{Experimental Settings}
\label{sec:exp_setting}

\textbf{Dataset.}
We conduct experiments on two datasets covering financial regulations and legal QA to evaluate  terminology understanding capabilities of \texttt{TermGPT} in high-stakes domains. The statistics of datasets are shown in Table \ref{tab:dataset_description}.

\textbf{(1) Financial Regulations Dataset.}
We construct a dataset from 425 regulatory rules extracted from officially published Chinese financial supervision documents, covering domains such as anti-money laundering, interbank lending, and loan management. Sentence graphs are built for all samples, and to prevent data leakage, we augment the nodes in the train set and test set separately.


\textbf{(2) JecQA Dataset.}
To further verify \texttt{TermGPT}'s generalization, we adopt JecQA \cite{zhong2020jec}, a large-scale legal QA dataset composed of knowledge-driven multiple-choice questions collected from the National Judicial Examination and related sources, which contains 26365 questions. For alignment with our regulatory dataset, we sample 1038 original questions from JecQA, using the same 7:3 split and data augmentation pipeline. 

We construct two test formats for both datasets: QCA-format, which assesses fine-grained semantic distinction, and QA-format, which evaluates answer generation and rule comprehension. All test sets are derived from QCA samples via LLM-based transformation. Details of data collection and processing are described in the Appendix A.
\begin{table}[t]  
    \setlength{\abovecaptionskip}{0.1cm}
    \centering
    \footnotesize
    \setlength{\tabcolsep}{1mm}
    \resizebox{\linewidth}{!}{
    \begin{tabular}{ccccccc} 
        \toprule   
        \multirow{2}{*}{Dataset} &  & \multicolumn{2}{c}{Original}
         & & \multicolumn{2}{c}{Augmented}\\
        \cmidrule{3-4}\cmidrule{6-7}
         & & Train & Test & & Train & Test \\
        \midrule
        \multirow{5}{*}{JecQA} & Economic law & 99 & 42 & & 802 & 560\\
        & Civil law & 268 & 114 & & 5586 & 4887\\
        & Civil Procedure Law & 211 & 90 & & 7120 & 6277 \\
        & Law of commerce & 150 & 64 & & 1112 & 657 \\ 
        \cmidrule{2-7}
        & Total & 728 & 310 & & 14620 & 12381 \\
        \midrule
        & Loan management & 99 & 29 & & 11718 & 5551 \\
        Financial & Plan fund management & 61 & 26 & & 12798 & 4043  \\
        Regulations & Risk management & 65 & 27 & & 7742 & 3589 \\
        & Company management & 83 & 35 & & 13974 & 6645 \\ 
        \cmidrule{2-7}
        & Total & 308 & 117 & & 46232 & 19828 \\
        \bottomrule  
    \end{tabular}}
    \caption{Statistics of the datasets.}
    \label{tab:dataset_description}
\end{table}

\begin{table*}[t]
\fontsize{0.26}{0.28}\selectfont
\setlength{\tabcolsep}{0.5pt}
\setlength{\heavyrulewidth}{0.12em}
\setlength{\lightrulewidth}{0.05em}
\setlength{\cmidrulewidth}{0.07em}
\setlength{\aboverulesep}{0.05pt}    
\setlength{\belowrulesep}{0.05pt}    
\resizebox{\textwidth}{!}{ 
\begin{tabular}{lc@{\hspace{0.5pt}}c@{\hspace{0.5pt}}c@{\hspace{0.5pt}}c@{\hspace{0.5pt}}c@{\hspace{0.5pt}}c@{\hspace{0.5pt}}c@{\hspace{0.5pt}}c@{\hspace{0.5pt}}c}
    \toprule
    \multirow{3}{*}{Dataset} & \multicolumn{4}{c}{JecQA} & & \multicolumn{4}{c}{Financial Regulations}\\
    \cmidrule{2-5}\cmidrule{7-10}
     & Accuracy & Precision &  Recall & F1 & & Accuracy & Precision &  Recall & F1 \\
    \midrule
    \multicolumn{10}{c}{\textit{Pre-trained Domain LLMs}} \\ 
    \midrule
    Lawyer-LlaMA(13B) & 0.480 	&	0.496 	&	0.480 	&	0.457 	&&	0.620 	&	0.621 	&	0.620 	&	0.620 	\\
    Xuanyuan(13B) & 0.808 & 0.812 & 0.808 & 0.807 && 0.701 & 0.707 & 0.701 & 0.701 \\
    \midrule
    \multicolumn{10}{c}{\textit{Contrastive Learning-based Methods}} \\ 
    \midrule
    AutoRegEmbed-Qwen3(8B) & 0.778 	&	0.779 	&	0.778 	&	0.778 	&&	0.883 	&	0.884 	&	0.883 	&	0.883 	\\
    PromptEOL-Qwen3(8B) &	0.831 	&	0.832 	&	0.831 	&	0.831 	&&	\underline{0.887} 	&	\underline{0.888} 	&	\underline{0.887} 	&	\underline{0.887} 	\\
    GritLM-Qwen3(8B) &	\underline{0.834} 	&	\underline{0.835} 	&	\underline{0.834} 	&	\underline{0.834} 	&&	0.876 	&	0.877 	&	0.876 	&	0.876 	\\
    NV-Embed-Qwen3(8B) &	0.816 	&	0.817 	&	0.816 	&	0.816 	&&	0.886 	&	0.887 	&	0.886 	&	0.886 	\\
    Flag-Embedding-Qwen3(8B) &	0.792 	&	0.793 	&	0.792 	&	0.792 	&&	0.879 	&	0.879 	&	0.879 	&	0.879 	\\
    \midrule
    \multicolumn{10}{c}{\textit{Commercial LLMs}} \\ 
    \midrule
    Qwen-2.5-plus (72B) & 0.894 	&	0.895 	&	0.894 	&	0.894 	&&	0.935 	&	0.936 	&	0.935 	&	0.936 	\\
    Deepseek-v3 (671B) & 0.884 	&	0.885 	&	0.884 	&	0.884 	&&	0.938 	&	0.939 	&	0.938 	&	0.938 	\\
    \midrule
    \multicolumn{10}{c}{\textit{Ours}} \\ 
    \midrule
    \texttt{TermGPT}-LlaMA (8B) &	 0.681	&	0.682	&	0.681	&	0.681	&&	0.856	&	0.857	&	0.857	&	0.857	\\
    \textbf{\texttt{TermGPT}-Qwen3 (8B)} &	\textbf{0.858} 	&	\textbf{0.858} 	&	\textbf{0.858} 	&	\textbf{0.858} 	&&	\textbf{0.908} 	&	\textbf{0.909} 	&	\textbf{0.908} 	&	\textbf{0.908} 	\\
    \midrule
    Improvement \raisebox{0.6ex}{\fontsize{0.26}{0.28}\selectfont 1} & +2.87\%	& +2.75\% &+2.87\%	&+2.87\%	&& +2.36\%	& +2.36\% &	+2.36\% & +2.36\% \\
    p-value \raisebox{0.6ex}{\fontsize{0.26}{0.28}\selectfont 2} & 0.008	& 0.008 & 0.008 & 0.008 && 0.001 & 0.001  & 0.001 	& 0.001 	\\
    \bottomrule
\end{tabular}
}
\caption*{
  \begin{minipage}[t]{\textwidth}
    \footnotesize\textsuperscript{1} Improvement over the best-performing baselines, which exclude commercial LLMs due to their substantially larger parameter scales.\\
    \footnotesize\textsuperscript{2} The improvement is significant based on a paired t-test at the significance level of 0.05 (p-value with paired t-test).
  \end{minipage}
}
\caption{Comparison of different models on terminology QCA task in terms of Accuracy, Precision, Recall and F1.}
\label{tab:overall_choice}
\end{table*}

\noindent \textbf{Comparison Methods.}
We compare \texttt{TermGPT} with three categories of baselines. The first includes pre-trained domain LLMs, such as Lawyer-LLaMA \cite{huang2023lawyer} and Xuanyuan \cite{zhang2023xuanyuan}, trained on legal and financial data to enhance domain-specific capabilities without task-specific supervision. The second consists of contrastive learning-based methods, including AutoRegEmbed \cite{deng2025following}, PromptEOL \cite{jiang2023scaling}, GritLM \cite{muennighoff2024generative}, NV-embed \cite{lee2024nv} and Flag-Embedding \cite{li2024making}, which optimize sentence embeddings via contrastive learning. The third includes commercial LLMs, such as Qwen-2.5-plus \cite{qwen2025qwen25technicalreport} and Deepseek-v3 \cite{liu2024deepseek}, evaluated via prompting to assess general language and domain knowledge.

\noindent \textbf{Metrics.} Following previous studies \cite{raffel2020exploring, lewis2019bart}, we adopt a set of generation metrics to evaluate the quality of the responses generated in the QA task, including BLEU-1, BLEU-4 \cite{papineni2002bleu}, ROUGE-1, ROUGE-L \cite{lin2004rouge}, METEOR \cite{banerjee2005meteor}, and BERTScore \cite{zhang2019bertscore}. These metrics collectively capture lexical overlap, fluency, and semantic similarity between the generated answers and the ground truth. For the QCA task, we report standard classification metrics: Accuracy, Precision, Recall, and F1, respectively.

\noindent \textbf{Implemental Details.} Model configuration and training setup are provided in the Appendix B.

\subsection{Overall Comparison (RQ1 \& RQ2)}

To address RQ1 and RQ2, we compare \texttt{TermGPT} with all baseline methods on terminology QCA (in Table \ref{tab:overall_choice}) and QA (in Table \ref{tab:overall_QA}) tasks. We highlight three main findings below.

\textbf{Result 1: Performance on Terminology QCA Tasks.} Contrastive learning-based methods slightly trail commercial LLMs but still outperform Lawyer-LLaMA by 63.00\% and Xuanyuan by 17.69\% on average. This highlights the efficiency of contrastive objectives in improving terminology-level discrimination, especially in low-parameter settings. Among contrastive learning-based methods, \texttt{TermGPT} outperforms the best baseline by 2.60\% on average. This is primarily because its multi-level supervision, which integrates sentence-level and token-level contrastive signals to better distinguish subtle semantic differences between terms.

\begin{figure}[t]
    \centering
    \includegraphics[width=1.0\linewidth]{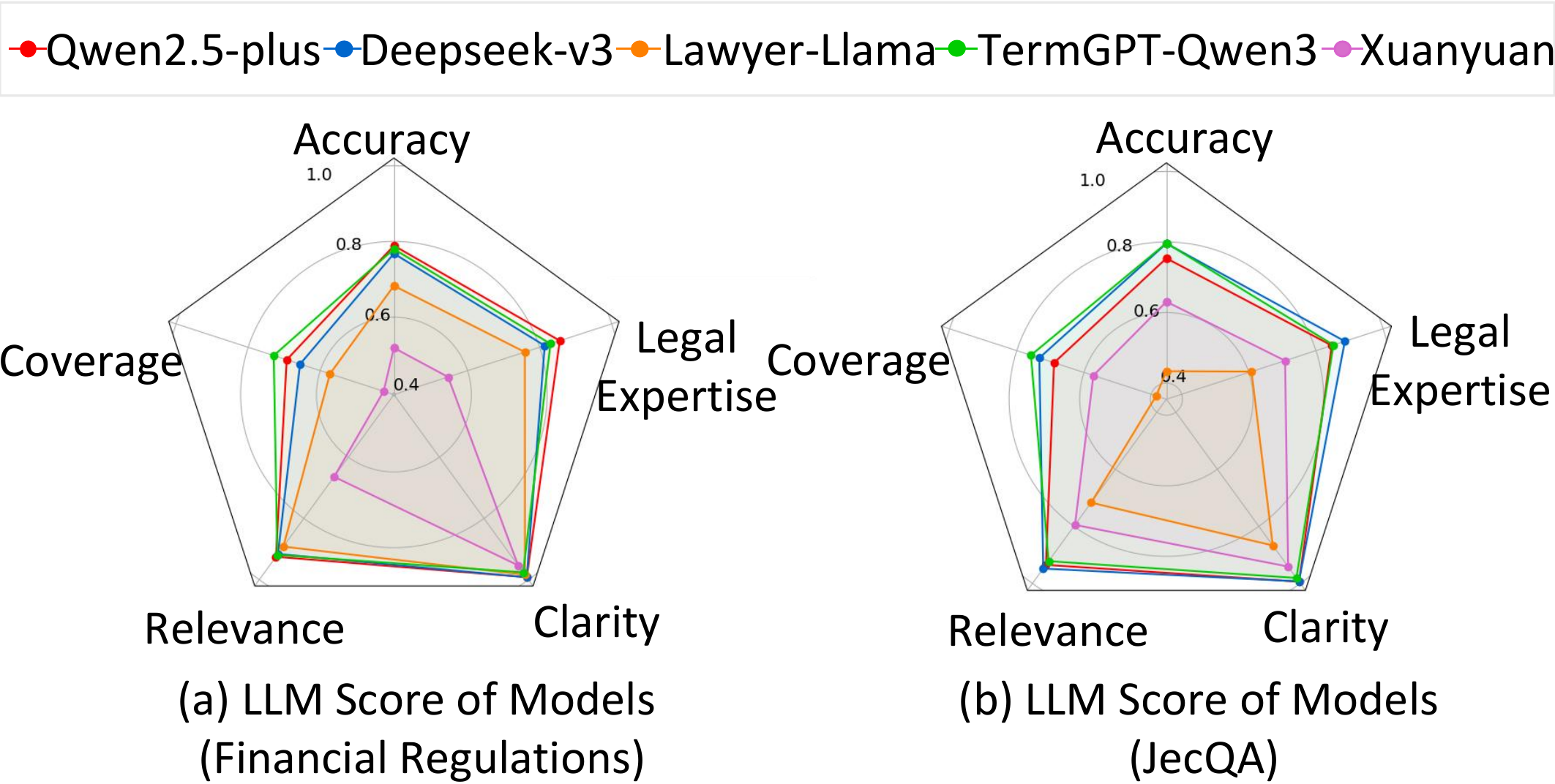}
    \caption{Comparison of different models on various datasets in terms of LLM Score.}
    \label{fig:radar}
\end{figure}

\textbf{Result 2: Performance on Terminology QA Tasks.} We evaluate the generated results with LLM in Figure~\ref{fig:radar}, which shows that \texttt{TermGPT} achieves comparable performance to commercial LLMs and significantly outperforms the domain-pretrained Lawyer-LLaMA and Xuanyuan. This suggests that although our model is smaller (8B), targeted supervision enables it to bridge the performance gap. Lawyer-LLaMA and Xuanyuan, while domain-pretrained, lacks task adaptation, limiting their effectiveness. Besides, the results in Table~\ref{tab:overall_QA} shows that \texttt{TermGPT} outperforms the best baseline by 6.14\% on average, benefiting from SFT, which improves task alignment, and contrastive learning, which enhances terminology-specific generation accuracy.

\textbf{Result 3: Backbone Selection.} We experiment with two backbone LLMs and observe that Qwen3 significantly outperforms LLaMA. In particular, TermGPT-Qwen3 yields an average performance gain of 15.98\% on the QCA task and 43.52\% on the QA task compared to TermGPT-LLaMA. Thus we adopt Qwen3 as the default backbone in our experiments.


\begin{table}[t]
\footnotesize
\resizebox{\linewidth}{!}{ 
\begin{tabular}{lccccc}
    \toprule
    \multicolumn{2}{l}{Dataset} & Accuracy & Precision &  Recall & F1 \\
    \midrule
    \texttt{TermGPT} & JecQA &   0.796 	&	0.797 	&	0.796 	&	0.796\\
    (w/o SFT)& Financial &	0.818 	&	0.819 	&	0.818 	&	0.818 	\\
    \cmidrule{2-6}
    \texttt{TermGPT}   &  JecQA &  0.834 	&	0.835 	&	0.834 	&	0.834\\
    (w/o CL)&   Financial &	   0.864 	&	0.865 	&	0.864 	&	0.864 	\\
    \cmidrule{2-6}
    \texttt{TermGPT} & JecQA   & 0.844 	&	0.844 	&	0.844 	&	0.844 	\\
    (w/o Token-level CL)&   Financial   &	0.899 	&	0.900 	&	0.899 	&	0.899 	\\
    \cmidrule{2-6}
    \texttt{TermGPT} & JecQA    &   0.849 	&	0.850 	&	0.849 	&	0.849 	\\
    (w/o Sentence-level CL)&   Financial   &	0.894 	&	0.895 	&	0.894 	&	0.894 	\\
    \cmidrule{2-6}
    \multirow{2}{*}{\texttt{TermGPT}} & JecQA    & 0.858 	&	0.858 	&	0.858 	&	0.858 	\\
    &   Financial   &	0.908 	&	0.909 	&	0.908 	&	0.908 	\\
    \bottomrule
\end{tabular}
}
\caption{Impact of CL and SFT on \texttt{TermGPT} performance.}
\label{tab:ablation_exp}
\end{table}
\begin{table*}[t]
\centering
\fontsize{2.2}{2.5}\selectfont
\setlength{\tabcolsep}{0.5pt}
\setlength{\heavyrulewidth}{0.15em}
\setlength{\lightrulewidth}{0.08em}
\setlength{\cmidrulewidth}{0.10em}
\setlength{\aboverulesep}{0.5pt}    
\setlength{\belowrulesep}{0.5pt}    
\resizebox{\textwidth}{!}{ 
\begin{tabular}{l c@{\hspace{0.7pt}}c@{\hspace{0.7pt}}c@{\hspace{0.7pt}}c@{\hspace{0.7pt}}c@{\hspace{0.7pt}}c c c@{\hspace{0.7pt}}c@{\hspace{0.7pt}}c@{\hspace{0.7pt}}c@{\hspace{0.7pt}}c@{\hspace{0.7pt}}c c}
\toprule
\multirow{3}{*}{Dataset} & \multicolumn{6}{c}{JecQA} & & \multicolumn{6}{c}{Financial Regulations}\\
\cmidrule(r{0.5pt}){2-7} \cmidrule(l{0.5pt}){9-14}
 & BLEU-1 & BLEU-4 & BERTscore  & ROUGE-1 & ROUGE-L & Meteor &  & BLEU-1 & BLEU-4 & BERTscore & ROUGE-1 & ROUGE-L & Meteor \\
\midrule
\multicolumn{14}{c}{\textit{Pre-trained Domain LLMs}} \\ 
\midrule
Lawyer-Llama (13B)	&	0.203 	&	0.105 &	0.678 	&	0.313 	&	0.232 	 	&	0.192 	&&	0.216 	&	0.096 	&	0.680 	&	0.310 	&	0.221 	&	0.176 	\\
Xuanyuan (13B)	&	0.260 	&	0.119 &	0.713	&	0.375 	&	0.311 	 	&	0.221 	&&	\underline{0.238} 	&	\underline{0.106} 	&	0.683 	&	\underline{0.334} 	&	0.254 	&	0.179 	\\
\midrule
\multicolumn{14}{c}{\textit{Contrastive Learning-based Methods}} \\ 
\midrule
AutoRegEmbed-Qwen3(8B)	&	\underline{0.261} 	&	0.121 &	\underline{0.714} 	&	\underline{0.375} 	&	0.313 	 	&	0.217 	&&	0.223 	&	0.081 	&	0.697 	&	0.318 	&	0.254 	&	0.185 	\\
PromptEOL-Qwen3(8B)	&	0.253 	&	0.119 &	0.712 	&	0.371 	&	0.309 	 	&	0.219 	&&	0.222 	&	0.082 	&	0.697 	&	0.316 	&	0.253 	&	0.188 	\\
GritLM-Qwen3(8B)	&	0.260 	&	\underline{0.122} &	0.714	&	0.374 	&	\underline{0.314} 	 	&	0.217 	&&	0.225 	&	0.084 	&	0.698 	&	0.316 	&	0.256 	&	0.186 	\\
NV-Embed-Qwen3(8B)	&	0.261 	&	0.121 &	0.713	&	0.375 	&	0.314 	 	&	\underline{0.220} 	&&	0.224 	&	0.080 	&	0.695 	&	0.317 	&	0.253 	&	0.187 	\\
Flag-Embedding-Qwen3(8B)	&	0.256 	&	0.120 &	0.714	&	0.372 	&	0.309 	 	&	0.217 	&&	0.229 	&	0.085 	&	\underline{0.699} 	&	0.324 	&	\underline{0.260} 	&	\underline{0.192} 	\\
\midrule
\multicolumn{14}{c}{\textit{Commercial LLMs}} \\ 
\midrule
Qwen-2.5-plus (72B)	&	0.160 	&	0.069 &	0.697	&	0.296 	&	0.218  	&	0.237 	&&	0.255 	&	0.111 	&	0.710 	&	0.361 	&	0.283 	&	0.213 	\\
Deepseek-v3 (671B)	&	0.261 	&	0.109 &	0.717	&	0.368 	&	0.299 	 	&	0.227 	&&	0.242 	&	0.084 	&	0.713 	&	0.333 	&	0.262 	&	0.221 	\\
\midrule
\multicolumn{14}{c}{\textit{Ours}} \\ 
\midrule
\texttt{TermGPT}-LlaMA (8B) & 0.152 & 0.074 & 0.624 & 0.251 & 0.182 & 0.132 && 0.212 & 0.091 & 0.656 & 0.297 & 0.217 & 0.152 \\
\textbf{\texttt{TermGPT}-Qwen3 (8B)} &	\textbf{0.266} 	&	\textbf{0.125} &	\textbf{0.716} 	&	\textbf{0.376} 	&	\textbf{0.316} 	 	&	\textbf{0.222} 	&&	\textbf{0.274} 	&	\textbf{0.126} 	&	\textbf{0.708} 	&	\textbf{0.371} 	&	\textbf{0.302} 	&	\textbf{0.201} 	\\
\midrule
Improvement \raisebox{0.6ex}{\fontsize{2.2}{2.5}\selectfont 1}  &	+1.92\% & +2.46\% & +0.28\% & +0.27\%& +0.64\% & +0.91\% && +15.13\% &	+18.87\% & +1.28\% & +11.08\% & 	+16.15\% & +4.69\%\\
p-value \raisebox{0.6ex}{\fontsize{2.2}{2.5}\selectfont 2} & 0.013	& 0.015 & 0.005	& 0.027 & 0.044  & 0.037 	&& 0.024 & 0.022 & 0.010 & 0.023 & 0.002 & 0.001 \\
\bottomrule
\end{tabular}
}
\caption*{
  \begin{minipage}[t]{\textwidth}
    \footnotesize\textsuperscript{1} Improvement over the best-performing baselines, which exclude commercial LLMs due to their substantially larger parameter scales.\\
    \footnotesize\textsuperscript{2} The improvement is significant based on a paired t-test at the significance level of 0.05 (p-value with paired t-test).
  \end{minipage}
}
\caption{Comparison of different models on terminology QA task in terms of BLEU, BERTscore, ROUGE and Meteor.}
\label{tab:overall_QA}
\end{table*}

\subsection{Ablation Experiment (RQ3)} To investigate RQ3, we conduct a series of ablation experiments to examine the individual contributions of multi-level contrastive learning and SFT to the performance of \texttt{TermGPT}. We report the results in Table~\ref{tab:ablation_exp}.

\textbf{Result 4: Effectiveness of Multi-level Contrastive Learning.} We first remove the token-level contrastive learning (\texttt{TermGPT} w/o Token-level CL)  and sentence-level contrastive learning (\texttt{TermGPT} w/o Sentence-level CL), respectively. The performance drops by 1.31\% and 1.28\% on average compared to the full multi-level version, indicating that aligning both global semantics and fine-grained terminology is critical for precise terminology understanding. Next, we compare the full model against two further variants: one without contrastive learning (\texttt{TermGPT} w/o CL) and one without SFT (\texttt{TermGPT} w/o SFT). The results show that removing contrastive learning leads to an average drop of 3.81\%, while removing SFT results in a drop of 8.55\%. This suggests that SFT helps the model better adapt to task-specific contexts, enhancing its ability to accurately understand and generate terminology, whereas contrastive learning improves the model’s capacity to distinguish between semantically similar terms. Together, these findings demonstrate that both SFT and multi-level contrastive learning are essential for maximizing terminology understanding in \texttt{TermGPT}.

\subsection{Performance Comparison in Different Domains (RQ4)}
We assess \texttt{TermGPT}’s performance across different sub-domains in both JecQA and financial regulations datasets. 

\begin{figure}[t]
    \centering
    \includegraphics[width=\linewidth]{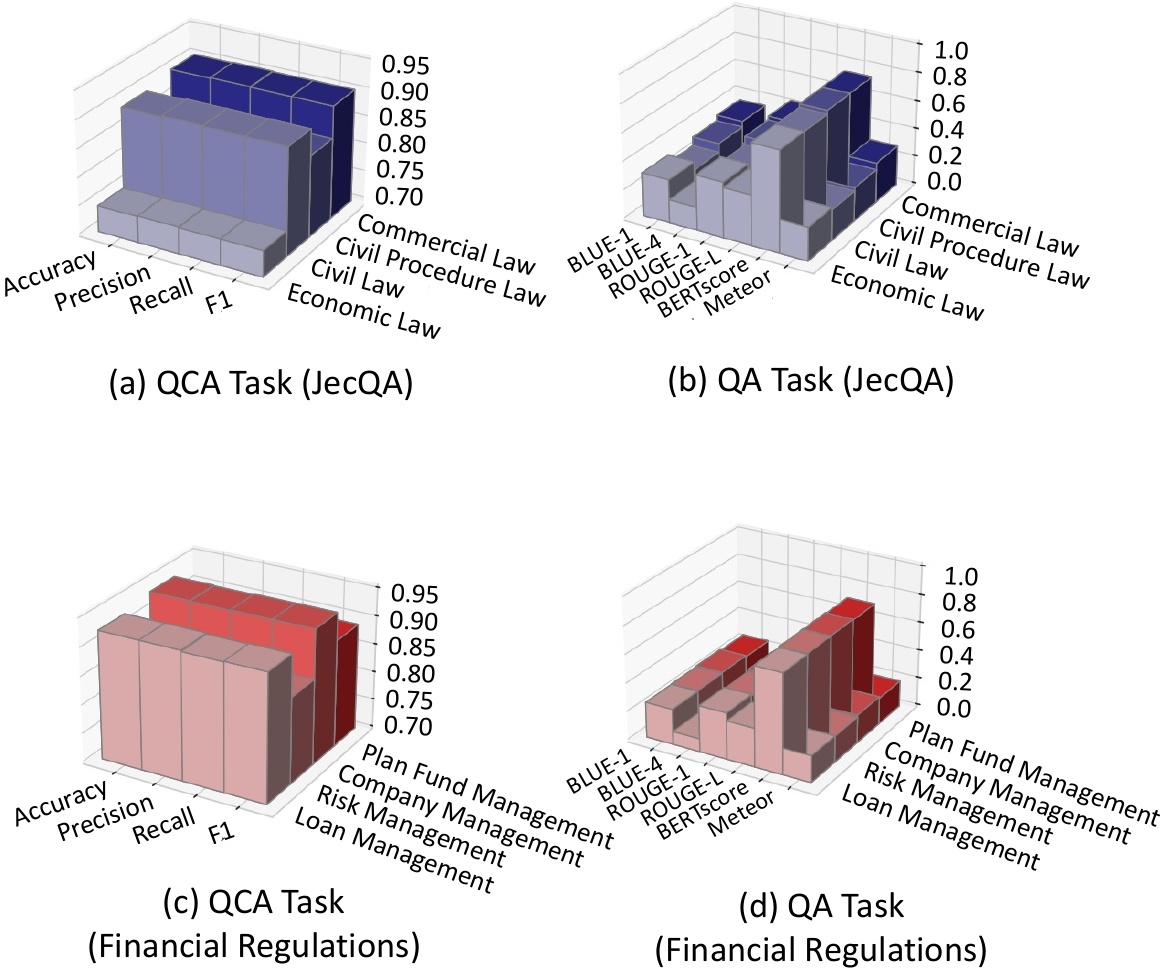}
    \caption{Performance of different domains on QCA and QA tasks.}
    \label{fig:catagories}
\end{figure}

\textbf{Result 5: Stable Terminology Understanding in Different Domains} As shown in Figure~\ref{fig:catagories}, the model maintains strong and balanced performance in both QCA and QA tasks in various domains. In  the QCA task, Economic Law and Risk Management show slightly lower scores, likely due to their limited training data. In the QA task, Civil Law lags slightly. This may be attributed to the broader scope and greater ambiguity of Civil Law questions, which pose a challenge for accurate modeling. These results highlight \texttt{TermGPT}’s robustness and reliability under diverse sub-domain conditions.

\section{Conclusions and Future Work}
In this paper, we propose \texttt{TermGPT}, a fine-tuning framework that integrates multi-level contrastive learning, and define the novel task of terminology-aware fine-tuning. By constructing sentence graphs to capture semantic relationships, we generate high-quality sample pairs to help LLMs better distinguish subtle terminology differences. Our \texttt{TermGPT} framework balances the optimization of global and local semantics, thereby improving sensitivity to terminological distinctions. Extensive experiments demonstrate that \texttt{TermGPT} effectively performs terminology-aware fine-tuning in various tasks and domain-specific applications.

An important direction for future work is enhancing the robustness of terminology-aware LLMs against adversarial inputs. In high-stakes domains, adversarial inputs may distort LLMs' interpretation of key terminology. Future work can explore adversarial training or robust optimization to improve reliability under such conditions.

\section{Acknowledgements}

This work was supported by the National Key R\&D Program of China (2024YFC3307702) and the cooperation project of ZJU-ZTCB Financial Technology Joint Research Center.

\bibliography{aaai2026}

\clearpage
\newpage

\appendix
\setcounter{secnumdepth}{1}
\section{Dataset Collection}
\label{appendix:dataset}

\subsection{Data Sources}

The \textbf{Financial Regulations Dataset} comprises 425 regulatory rules manually extracted from a diverse range of officially issued Chinese financial regulatory documents. These regulations cover key areas such as anti-money laundering, interbank lending, loan management, and risk control. The source documents were collected from publicly accessible and authoritative regulatory sources, including:

\begin{itemize}
    \item Company Management:
    \begin{itemize}
    \item \textit{Company Law of the People's Republic of China (2018 Amendment)}
    \item \textit{Law of the People's Republic of China on Commercial Banks (2015 Amendment)}
    \item \textit{Law of the People's Republic of China on Banking Regulation and Supervision (2006 Amendment)}
    \item \textit{Notice of the General Office of the CBRC on Strengthening the Qualification Examination of Major Shareholders of Small and Medium-sized Commercial Banks}
    \item \textit{Guidelines on the System of Independent Directors and External Directors of Joint-Stock Commercial Banks}
    \item \textit{Trial Measures for the Performance Evaluation of Directors of Commercial Banks}
    \item \textit{Guidelines for the Work of Boards of Directors of Commercial Banks}
    \item \textit{Corporate Governance Guidelines for Commercial Banks}
    \item \textit{Provisional Regulations on the Investment of Financial Institutions in Equities}
    \item \textit{Provisional Measures for the Administration of Equity of Commercial Banks}
    \item \textit{Notice on the Implementation of the Provisional Measures for the Administration of Equity of Commercial Banks}
    \item \textit{Notice on Regulating Matters Concerning the Equities of Commercial Banks}
    \item \textit{Notice on Issues Concerning the Investment and Financing of Financial Enterprises in Local Governments and State-owned Enterprises}
    \item \textit{Guiding Opinions on Strengthening the Supervision of Equity Investment by Financial Institutions}
    \item \textit{Measures for the Administration of Qualifications of Directors (Supervisors) and Senior Executives of Banking Financial Institutions}
    \item \textit{Implementation Measures for the Administrative Licensing Items of Township and Village Banks and Other Small and Medium-sized Financial Institutions}
    \item \textit{Guiding Opinions on Strengthening the Rule of Law in the Work of Banking Financial Institutions}
    \item \textit{Implementation Measures for the Administrative Licensing Items of Domestic Commercial Banks (2018 Revision)}
    \item \textit{Measures for the Supervision and Administration of the Equity of Listed Companies}
    \item \textit{Code of Corporate Governance for Listed Companies}
\end{itemize}

    \item Risk Management:
    \begin{itemize}
    \item \textit{Administrative Measures for Financial Institutions to Report Suspicious Transactions Related to Terrorist Financing}
    \item \textit{Administrative Measures for Customer Identification and Preservation of Customer Identity Information and Transaction Records by Financial Institutions}
    \item \textit{Trial Measures for the Management of Information Submission by Anti-Money Laundering Reporting Entities}
    \item \textit{Trial Procedures for the Electronic Submission of Large-Value and Suspicious Transaction Reports via Internet}
    \item \textit{Notice on Further Strengthening Anti-Money Laundering Work for Cross-border RMB Business}
    \item \textit{Notice on Strengthening Anti-Money Laundering Cooperation in Cross-border Business of Financial Institutions}
    \item \textit{Guidelines for the Evaluation and Classification of Money Laundering and Terrorist Financing Risks by Financial Institutions}
    \item \textit{Trial Measures for the Supervision and Administration of Anti-Money Laundering Work of Financial Institutions}
    \item \textit{Notice on Strengthening the Reporting of Suspicious Transactions Involving Persons on the Sanctions List}
    \item \textit{Notice on Strengthening the Management of Payment System Liquidity Risk and System Operation Risk}
    \item \textit{Guiding Opinions on the Implementation of New Supervision Standards by Chinese Commercial Banks}
    \item \textit{Measures for the Administration of Liquidity Risk of Commercial Banks}
    \item \textit{Notice on Printing and Distributing the Information Disclosure Format for Liquidity Coverage Ratio of Commercial Banks}
    \item \textit{Measures for the Disclosure of Net Stable Funding Ratio of Commercial Banks}
\end{itemize}

    \item Plan Fund Management:
    \begin{itemize}
    \item \textit{Law of the People's Republic of China on Commercial Banks (2015 Amendment)}
    \item \textit{Measures for the Punishment of Financial Violations}
    \item \textit{Guiding Opinions on the Implementation of New Supervision Standards by Chinese Commercial Banks}
    \item \textit{Trial Measures for the Administration of Capital of Commercial Banks}
    \item \textit{Measures for the Administration of Capital of Commercial Banks (Trial)}
    \item \textit{Corporate Governance Guidelines for Commercial Banks}
    \item \textit{Notice on Issuing Policy Documents for Capital Supervision and Allocation of Commercial Banks}
    \item \textit{Measures for the Administration of Leverage Ratio of Commercial Banks (2015)}
    \item \textit{Opinions on Further Supporting Capital Tools Innovation in Commercial Banks}
    \item \textit{Announcement No. 3 [2018] of the People's Bank of China — Matters Concerning the Issuance of Capital Instruments by Financial Institutions}
\end{itemize}

\item Loan Management:
\begin{itemize}
    \item \textit{Law of the People's Republic of China on Commercial Banks (2015 Amendment)}
    \item \textit{Rules on Loans}
    \item \textit{Measures for the Punishment of Financial Violations}
    \item \textit{Guidelines for Loan Risk Classification}
    \item \textit{Notice on Strengthening Credit Risk Management for Commercial Banks' Personal Consumption Loans}
    \item \textit{Notice on Regulating Credit Risk Management for Personal Loans}
    \item \textit{Guidelines for Credit Risk Management of Real Estate Loans}
    \item \textit{Asset Management Guidelines for Commercial Banks}
    \item \textit{Notice on Strengthening Macroeconomic Management and Regulating Credit Structure}
    \item \textit{Notice on Further Strengthening Macroeconomic Management and Regulating Credit Structure}
    \item \textit{Guidelines for Loan Risk Classification}
    \item \textit{Notice on Preventing and Controlling High-risk Loans in Certain Industries}
    \item \textit{Notice on Improving Credit Risk Early Warning Mechanisms}
    \item \textit{Notice on Further Strengthening the Implementation of Credit Policies in High-risk Industries}
\end{itemize}

\end{itemize}

These documents were sourced from authoritative public channels, including ministries and commissions such as the China Banking and Insurance Regulatory Commission (CBIRC), the People’s Bank of China (PBOC), the State Council, as well as their official websites, legal bulletin platforms, and central bank regulatory announcements.

\subsection{Annotation Process}

Each regulatory rule was segmented into sentence-level samples and annotated with domain-specific terminology. To ensure annotation accuracy and consistency, term identification was conducted by a panel of experts in financial law and regulatory compliance. A rigorous annotation process was followed, incorporating cross-validation and consensus-based resolution protocols to ensure high-quality results.

\subsection{Data Augmentation}
The prompts used for token-level and sentence-level data augmentation are shown in Table~\ref{tab:token-prompt} and Table~\ref{tab:sentence-prompt}, respectively. To generate high-quality augmentation samples, we leverage \textbf{Qwen-2.5-plus}, a powerful open-source LLM known for its strong performance in both general and domain-specific understanding. 

The QCA instances for token-level ambiguity discrimination are shown in Table~\ref{tab:qca-examples-token} and those for sentence-level ambiguity discrimination are shown in Table~\ref{tab:qca-examples-sentence}.

\begin{table*}[h]
\centering
\tiny
\renewcommand{\arraystretch}{1.3}
\begin{tabular}{|p{0.95\textwidth}|}
\hline
\textbf{Token-Level Data Augmentation Prompt} \\
\hline
\begin{minipage}[t]{0.95\textwidth}
\raggedright
Please generate a question-answer pair based on the provided sentence that contains a specific term and its similar words. Follow the instructions below:

\begin{enumerate}
    \item You must generate the question based on the background sentence, and the question should focus on the specified term \texttt{\{anchor\}}. Do not copy the background sentence directly.
    \item \texttt{\{anchor\}} must be the correct answer.
    \item After generating the question, convert it into a single declarative sentence (rephrased sentence) that answers the question directly. Do not add any information beyond what is contained in the question itself, and do not use content from the background sentence that is not relevant.
    \item Only return the required fields below. Do not include any explanations or extra content.
    \item Avoid vague interrogatives such as ``Which of the following...'' or ``The following is/are...''.
    \item You must strictly follow the output format below. Any additional content will cause parsing errors.
\end{enumerate}

Background sentence (containing the specified term):\\
\texttt{\{anchor\}} \\[0.5em]
Specified term (correct answer):\\
\texttt{\{anchor\}} \\[0.5em]

Please generate the output in the following format:\\
\texttt{<Question>: ... ...}\\
\texttt{<Correct Answer>: \{anchor\}}\\
\texttt{<Rephrased Sentence>: ... ...}\\
\texttt{ }
\end{minipage}
\\
\hline
\end{tabular}
\caption{Prompt used for token-level data augmentation in in QCA task construction.}
\label{tab:token-prompt}
\end{table*}

\begin{table*}[h]
\centering
\tiny
\renewcommand{\arraystretch}{1.3}
\begin{tabular}{|p{0.95\textwidth}|}
\hline
\textbf{Sentence-Level Data Augmentation Prompt} \\
\hline
\begin{minipage}[t]{0.95\textwidth}
\raggedright
Please generate a multiple-choice question in the Question-Choice-Answer (QCA) format based on the provided anchor sample and negative sample. The requirements are as follows:

\begin{enumerate}
    \item The question must contain one question (Question), four choices (Choices), and a correct answer (Answer).
    \item Choice A should be the correct option, derived from the anchor sample. Choice B should be an incorrect option derived from the negative sample. Choices C and D are incorrect options generated by you.
    \item All four choices must be complete sentences, not words or phrases.
    \item Choice B should be a wrong answer based on the content of the negative sample.
    \item Choices C and D should be hard negative options for A, incorporating elements from the negative sample while maintaining logical and grammatical coherence.
    \item The correct answer should be presented as the full sentence, not labeled by letters such as A/B/C/D.
    \item The question must avoid vague formulations such as ``Which of the following...'' or ``Which of the following is/are not...''. The correct answer must not contain template-like phrases such as ``According to... regulations''.
    \item Please strictly follow the output format below (including angle brackets), and do not include any explanation or annotation.
\end{enumerate}

Anchor Sample (correct information source):\\
\texttt{\{anchor\}} \\[0.5em]
Negative Sample (incorrect information source):\\
\texttt{\{negative\}} \\[0.5em]

Please generate the question in the following format:\\
\texttt{<Question>: ... ...}\\
\texttt{<Choice A>: ... ...}\\
\texttt{<Choice B>: ... ...}\\
\texttt{<Choice C>: ... ...}\\
\texttt{<Choice D>: ... ...}\\
\texttt{<Correct Answer>: ... ...} \\
\texttt{ }
\end{minipage}
\\
\hline
\end{tabular}
\caption{Prompt used for sentence-level data augmentation in QCA task construction.}
\label{tab:sentence-prompt}
\end{table*}

\subsection{Data Access}

The dataset will be publicly released upon publication of the paper. 
A permanent download link and license terms will be made available on the project website. We commit to ensuring long-term accessibility for academic use.

\begin{table*}[htbp]
\centering

\begin{tabular}{@{}p{0.31\linewidth} p{0.31\linewidth} p{0.31\linewidth}@{}}
\toprule
\multicolumn{3}{@{}l}{\textbf{Token-level QCA Example 1}} \\
\midrule
\multicolumn{3}{@{}p{0.93\linewidth}}{
\textbf{Question:} Among the causes that interrupt the statute of limitations, to whom must the agent of the obligee make a request for it to constitute a valid cause of interruption?
} \\
\multicolumn{3}{@{}p{0.93\linewidth}}{\textbf{Answer: Debtor}} \\
\textbf{Choice 1:} Guarantor  & \textbf{Choice 2:} Policyholder & \textbf{ \checkmark Choice 3:} Debtor  \\
\textbf{Choice 4:} Joint creditor & \textbf{Choice 5:} Property owner & \textbf{Choice 6:} Sub-debtor \\
\textbf{Choice 7:} Mortgagee & \textbf{Choice 8:} Unsecured creditor & \\
\bottomrule
\end{tabular}

\vspace{0.8em}

\begin{tabular}{@{}p{0.31\linewidth} p{0.31\linewidth} p{0.31\linewidth}@{}}
\toprule
\multicolumn{3}{@{}l}{\textbf{Token-level QCA Example 2}} \\
\midrule
\multicolumn{3}{@{}p{0.93\linewidth}}{
\textbf{Question:} According to the \textit{Guidelines on Corporate Governance of Commercial Banks}, if a person fails to attend relevant meetings in person twice in a row and also fails to entrust another person to attend on their behalf, or fails to attend at least two-thirds of the meetings in person in a year, they shall be deemed unable to perform their duties. What is the term used to describe such personnel?
} \\
\multicolumn{3}{@{}p{0.93\linewidth}}{\textbf{Answer: Supervisor}} \\
\textbf{Choice 1:} Director & \textbf{Choice 2:} External supervisor & \textbf{Choice 3:} Executive oversight \\
\textbf{Choice 4:} Lead regulator & \textbf{\checkmark Choice 5:} Supervisor & \textbf{Choice 6:} Board of supervisors \\
\bottomrule
\end{tabular}

\vspace{0.8em}

\begin{tabular}{@{}p{0.31\linewidth} p{0.31\linewidth} p{0.31\linewidth}@{}}
\toprule
\multicolumn{3}{@{}l}{\textbf{Token-level QCA Example 3}} \\
\midrule
\multicolumn{3}{@{}p{0.93\linewidth}}{
\textbf{Question:} In the process of equity management of commercial banks, if a violation occurs causing serious consequences, in addition to penalizing the chairman of the board, which other position’s responsible personnel may also receive warnings, fines, or disqualification by the CBRC or its local offices?
} \\
\multicolumn{3}{@{}p{0.93\linewidth}}{\textbf{Answer: Board secretary}} \\
\textbf{\checkmark Choice 1:} Board secretary & \textbf{Choice 2:} Director & \textbf{Choice 3:} Board of directors \\
\bottomrule
\end{tabular}

\caption{Illustrative QCA examples for token-level ambiguity discrimination.}
\label{tab:qca-examples-token}
\end{table*}

\begin{table*}[htbp]
\centering

\begin{tabular}{@{}p{0.96\linewidth}@{}}
\toprule
\textbf{Sentence-level QCA Example 1} \\
\midrule
\textbf{Question:} What procedural requirements should listed companies follow when implementing phased equity incentive plan? \\
\textbf{Choice 1:} If the company lowers the exercise or grant price in a modified plan, it only needs approval from the supervisory board without further review by the general meeting. \\
\textbf{Choice 2:} Before the equity incentive plan is approved by the general meeting, the listed company may change it at will with board approval only, without following any additional procedures. \\
\textbf{Choice 3:} After the plan is approved by the general meeting, changes involving accelerated vesting or early release of restrictions may be made solely with board approval. \\
\textbf{\checkmark Choice 4:} Before each grant, the board of directors shall be convened to determine the grant price and exercise arrangement based on the incentive plan and principles set during the initial grant. If the grant conditions are not met, no rights shall be granted, and ungranted rights shall not be deferred. \\
\bottomrule
\end{tabular}

\begin{tabular}{@{}p{0.96\linewidth}@{}}
\toprule
\textbf{Sentence-level QCA Example 2} \\
\midrule
\textbf{Question:} When the scope of a guarantee is not agreed upon or unclear between parties, how should the guarantor be liable? \\
\textbf{\checkmark Choice 1:} The guarantor shall be liable for the entire debt.   \\
\textbf{Choice 2:} The guarantor is only liable for part of the debt if a ruling has been received within six months. \\
\textbf{Choice 3:} The guarantor assumes liability only after applying for confirmation of liability at the intermediate people's court at the arbitration commission’s location. \\
\textbf{Choice 4:} The guarantor is only liable if there is evidence that the debt is subject to cancellation. \\
\bottomrule
\end{tabular}

\begin{tabular}{@{}p{0.96\linewidth}@{}}
\toprule
\textbf{Sentence-level QCA Example 3} \\
\midrule
\textbf{Question:} What types of first-instance civil cases fall under the jurisdiction of the Supreme People's Court? \\
\textbf{Choice 1:} Retrial cases and first-instance cases accepted by the intermediate people’s court. \\
\textbf{\checkmark Choice 2:} Cases that the court deems should be tried by itself, as well as cases that have significant national impact.  \\
\textbf{Choice 3:} Cases with national impact and where the defendant's whereabouts are unknown at the time of filing. \\
\textbf{Choice 4:} Cases the court deems should be tried by itself and first-instance cases accepted by the intermediate court. \\
\bottomrule
\end{tabular}

\caption{Illustrative QCA examples for sentence-level ambiguity discrimination.}
\label{tab:qca-examples-sentence}
\end{table*}

\section{Implemental Details}
\subsection{Model Configuration}

Our experiments leverage two state-of-the-art instruction-tuned language models as the backbone encoders: Qwen3-8B-Instruct and LLaMA3-8B-Instruct.

Qwen3-8B-Instruct is a large-scale generative language model developed for advanced natural language generation tasks. It is optimized for instruction-following scenarios and capable of producing high-quality semantic representations.

LLaMA3-8B-Instruct, released by Meta, is another leading instruction-tuned model with enhanced capabilities in contextual understanding, multi-task learning, and semantic reasoning.

Both models are pre-trained on large-scale corpora and fine-tuned using instruction-following objectives. In our setup, we extract both sentence-level and entity-level embeddings by retrieving the hidden state of the final token in the output sequence, which effectively captures the semantic information of the entire input span.

These embeddings are then used in downstream tasks such as sentence classification and entity-level representation learning.

\subsection{Training Details}

During supervised fine-tuning (SFT), we employ LoRA with rank 16 and $\alpha$ 32 to reduce trainable parameters and computational cost. To further optimize training under limited resources, we adopt DeepSpeed-ZeRO2 for partitioning optimizer states and gradients, combined with bf16 precision for improved efficiency. Optimization is conducted using the AdamW optimizer with linearly decaying learning rate scheduler, where the learning rate is tuned in the range of [5e-6, 1e-5, 5e-5]. The batch size is set to 16, and each epoch includes 3 full passes over the training data. We use cross-entropy loss for supervised objectives and contrastive loss with temperature values tuned in [0.05, 0.1]. All experiments are performed on two NVIDIA A100 (40GB) GPUs.

\end{document}